\documentclass[11pt]{article}

\usepackage[preprint]{acl}

\usepackage{times}
\usepackage{latexsym}
\usepackage{lscape} 
\usepackage{makecell}

\usepackage[T1]{fontenc}
\usepackage[utf8]{inputenc}

\usepackage{microtype}

\usepackage{inconsolata}

\usepackage{graphicx}
\usepackage{todonotes}
\usepackage{pifont}
\usepackage{booktabs}
\usepackage{csquotes}
\usepackage{hyperref}
\usepackage{float}
\usepackage{amssymb}
\usepackage{multirow}
\usepackage{multicol} % Added for multicols environment
\newcommand{\cmark}{\ding{51}}%
\newcommand{\xmark}{\ding{55}}%

\usepackage{graphicx,wrapfig}

\usepackage{adjustbox}

\title{Arctic-Extract Technical Report}

\author{Mateusz Chiliński, Julita Ołtusek, and Wojciech Jaśkowski \\
  \\Snowflake AI Research \\
  \texttt{name.surname@snowflake.com} \\}

\begin{document}
\maketitle
\begin{abstract}
Arctic-Extract is a state-of-the-art model designed for extracting structural data (question answering, entities and tables) from scanned or digital-born business documents. Despite its SoTA capabilities, the model is deployable on resource-constrained hardware, weighting only 6.6 GiB, making it suitable for deployment on devices with limited resources, such as A10 GPUs with 24 GB of memory. Arctic-Extract can process up to 125 A4 pages on those GPUs, making suitable for long document processing. This paper highlights Arctic-Extract's training protocols and evaluation results, demonstrating its strong performance in document understanding.
\end{abstract}

\section{Introduction}
The field of Document Processing has witnessed remarkable advancements with the advent of multimodal large language models (MLLMs). These models address the inherent challenges of processing documents where spatial information is critical - a task that traditional large language models (LLMs) often struggle to handle effectively. However, many state-of-the-art MLLMs remain prohibitively expensive for enterprise-grade applications, limiting their practical adoption.

Our research is providing a successor to Arctic-TILT model \cite{borchmann2024arctictiltbusinessdocumentunderstanding}, aiming to surpass its performance while adhering to similar cost and hardware constraints. The primary objective of Arctic-Extract was to harness the capabilities of modern MLLMs to achieve advanced document understanding. 

\section{Related work}
The introduction of the Vision Transformer (ViT) \cite{dosovitskiy2021imageworth16x16words} marked a paradigm shift in computer vision, which subsequently influenced the development of MLLMs. By enabling the processing of images in a manner analogous to text, ViTs paved the way for their integration into existing LLM architectures. A significant milestone in this evolution was the Florence model \cite{yuan2021florencenewfoundationmodel} and Donut\cite{kim2022ocrfreedocumentunderstandingtransformer}, which laid the groundwork for subsequent advancements. More recently, models such as LLaVA \cite{liu2023visualinstructiontuning}, Qwen-VL \cite{bai2023qwenvlversatilevisionlanguagemodel}, Pixtral \cite{agrawal2024pixtral12b}, and DeepSeek \cite{wu2024deepseekvl2mixtureofexpertsvisionlanguagemodels} have further expanded the field. While these models demonstrate impressive capabilities, their high computational costs often render them unsuitable for enterprise-grade document processing. Moreover, their smaller variants frequently fail to deliver adequate extraction accuracy, as they are designed for general-purpose tasks rather than specialized document processing.

In this work, we introduce Arctic-Extract, an enterprise-grade, cost-effective solution tailored for advanced document understanding. Arctic-Extract is capable of processing documents of up to 125 pages on a single A10 GPU (24GB memory), delivering exceptional performance across both public and internal datasets. Remarkably, it matches or surpasses the performance of state-of-the-art models with over 400 billion parameters, all while being significantly more affordable and resource-efficient.

\section{Arctic-Extract}

\begin{table*}[htbp]
\centering
\caption{Comprehensive model performance comparison across different evaluation categories.}
\label{tab:comprehensive_performance}
\begin{adjustbox}{center}
\renewcommand{\arraystretch}{1.5} % Adjust row spacing
\begin{tabular}{@{}lcccc@{}}
\toprule
\multirow{2}{*}{\textbf{Model}} & \multicolumn{3}{c}{\textbf{QA}} & \multirow{2}{*}{\textbf{Table Extraction}} \\
\cmidrule(lr){2-4}
 & \textbf{Visual} & \textbf{Text} & \textbf{Multilingual Text} & \\ 
\midrule
\textbf{Arctic-Extract} & \textbf{64.2} & 78.94 & \textbf{70.7} & 72.0 \\
Arctic-TILT & 44.4 & -- &  36.3 & \textbf{72.5} \\
Claude 4 Sonnet & 51.3 & 79.65 &  65.1 &70.7 \\
GPT5 & 58.5 & 74.27 &  66.8 & 65.1 \\
LLama 3.1 405B & -- & \textbf{80.37} & 67.0 & -- \\
LLama 3.1 8B & -- & 59.13 & 57.1 & -- \\
LLama 4 Scout & -- & 74.30 & 63.8 & -- \\
Mistral 7B & -- & 13.50 & 47.9 & -- \\
Pixtral 12B & 35.9 & 55.23 & 57.8 & 16.0 \\
Qwen 2.5-VL & 50.8 & 66.76 & 63.3 & 56.0 \\
\bottomrule
\end{tabular}
\end{adjustbox}
\end{table*}

We decided to use the architecture of Qwen2.5-VL \cite{bai2025qwen25vltechnicalreport} because of its unique property that is the merging of the 4 tokens into one vision token. This allows standard A4 page to be compressed to around 1000 tokens with high accuracy, and thus allowing to fit more content into the context window. The context window of the model is 128 000 tokens, and is able to process 125 pages at once, which is essential for processing large documents in the real world.

\subsection{Capabilities}

Arctic-Extract exhibits the ability to:
\begin{itemize}
  \itemsep0em 
  \item Perform question answering (QA), including QA from images.
  \item Extract entities such as invoice numbers.
  \item Extract tables (refer to Appendix ~\ref{business_table_example} and ~\ref{te_challenges} for more details and examples).
\end{itemize}

These functionalities are applicable to both digitally-born and scanned documents, including those containing handwritten content.

We support the following languages: Arabic, Bengali, Burmese, Cebuano, Chinese, Czech, Dutch, English, French, German, Hebrew, Hindi, Indonesian, Italian, Japanese, Khmer, Korean, Lao, Malay, Persian, Polish, Portuguese, Russian, Spanish, Tagalog, Thai, Turkish, Urdu, Vietnamese.

\subsection{Training protocol}
To fine-tune the model, we utilized the ModelScope Swift Trainer \cite{zhao2024swiftascalablelightweightinfrastructure}. The training was conducted on 8 NVIDIA H200 GPUs, each with 141 GB of memory. The batch size was set to 1 per GPU, with gradient accumulation steps of 16. The model was trained for 1 epoch using a learning rate of $1 \times 10^{-4}$ and a weight decay of 0.01. A cosine learning rate scheduler was employed. The training process, which utilized bf16 precision and the AdamW optimizer \cite{loshchilov2019decoupledweightdecayregularization}, was completed in approximately 4 days.

For parameter-efficient fine-tuning, we adopted LoRA \cite{hu2021loralowrankadaptationlarge}, as it demonstrated superior performance compared to full-parameter fine-tuning in our use case. While memory constraints were not a primary concern, the challenges associated with long-context training were significant. LoRA was configured with a rank of 32 and an alpha of 32. The fine-tuning was applied to linear layers, identified using the following regular expression:
\begin{displayquote}
\footnotesize
\texttt{\^{ }(model)(?!.*(lora\_A|lora\_B|base\_layer|\\emb|wte|shared|lm\_head|output|score|v\_head|\\classifier)).*}
\end{displayquote}

The LoRA obtained during training was merged with the base model using the ModelScope Swift framework. To ensure the entire model fits within the 24GB memory of an A10 GPU, we applied 4-bit quantization using AWQ \cite{lin2024awqactivationawareweightquantization}. The quantization process was carried out with the AutoAWQ implementation \cite{githubGitHubCasperhansenAutoAWQ}, utilizing a group size of 128 and the GEMM quantization type. The final model, when loaded with vLLM \cite{kwon2023efficient}, occupies only 6.6 GiB of memory. That allows it to process up to 125 A4 pages (digital-born or scanned) on a single A10 GPU with 24 GB of memory, making it suitable for long document processing, at a very competitive cost.

\subsection{Training data}

We have used a total of 372,544 datapoints for the training, spanning across 35 datasets, out of which 13 covered the Table Extraction problem (33,916 datapoints), 13 covered the QA problem (112,901 datapoints), and the remaining 9 covered multilingual problems (225,735 datapoints). Most of the datasets were internal ones; however, the publicly available datasets (or datasets derived/processed based on them) include TABFACT \cite{2019TabFactA}, DeepForm \cite{borchmann2021due}, Kleister NDA \cite{Stanis_awek_2021}, Kleister Charity \cite{Stanis_awek_2021}, DocVQA \cite{mathew2021docvqadatasetvqadocument}, DUDE \cite{vanlandeghem2023documentunderstandingdatasetevaluation}, InfographicVQA \cite{mathew2021infographicvqa}, and TAT-DQA \cite{Zhu_2022}. The public datasets used for composition of our multilingual datasets were GermanQuAD \cite{möller2021germanquad}, SQAC \cite{maria}, frenchQA \cite{frenchQA2023}, SQuAD-it \cite{10.1007/978-3-030-03840-3_29}, PoQuAD \cite{10.1145/3587259.3627548}, BR-QuAD-2.0 \cite{githubGitHubPiEspositobrquad20}, IDSEM \cite{sanchez2022idsem}, SQuAD-es v2.0 \cite{carrino2019automatic}, and InvoicesReceiptsPT \cite{cruz_2022_6371710}.

To address the Table Extraction (TE) problem, dedicated datasets were developed, as this represents a novel research task for which no publicly available datasets currently exist. The identified TE problem differs fundamentally from Table Recognition \cite{Zanibbi2004}, which is primarily concerned with detecting and extracting tabular structures from documents. In contrast, TE focuses on the transformation of information contained within unstructured documents into structured tabular formats. The datasets were purpose-built to align with the specific objectives and challenges of the TE task. Their development followed an iterative process involving manual annotation with systematic error analysis and consistency evaluation within a feedback loop, complemented by automated conversion from nested data representations to tabular formats. Additionally, synthetic data generation was employed to create artificial table layouts, specifically designed to address task-specific challenges that were underrepresented or absent in other data sources.

To give the research community a grasp of the datasets used as the internal ones, we refer the reader to Arctic-TILT paper, as they partially overlap \cite{borchmann2024arctictiltbusinessdocumentunderstanding}.

\section{Experiments}

To comprehensively evaluate our new model, we conducted extensive experiments across four main categories: visual document understanding, multilingual text processing, English text comprehension, and table extraction. We aggregated results from all experiments into a unified comparison table (Table \ref{tab:comprehensive_performance}) to provide a comprehensive overview of model performance across different tasks and datasets. Each category is described in detail below, with complete experimental results presented in the appendix.

Our experimental setup was constrained by model availability and capabilities. Some models could not be evaluated on all tasks due to limitations in supported functionality or languages. For instance, Mistral 7B and LLama models lack image input support, precluding their evaluation on visual tasks (which in our case, are QA visual evaluation and table extraction). While API-based models (GPT5 and Claude 4 Sonnet) support visual tasks, the proprietary nature of our datasets required evaluation through the Cortex AI platform, where larger file sizes frequently resulted in processing errors. In the aggregated table, we present results excluding datapoints that could not be processed by all models, ensuring fair comparison across all evaluated systems.

However, given that Arctic-Extract is specifically designed to address input size limitations, we also provide complete experimental results in the appendix, including cases where certain models failed to process inputs entirely (receiving a score of 0 for such instances). All comparisons utilize ANLS* \cite{peer2025anlsuniversaldocument} as the primary evaluation metric, as it is most appropriate for document question answering tasks.

\subsection{English text evaluation}
Looking at the SQuAD2.0 evaluation table (Table \ref{tab:model_comparison}), it presents a performance comparison of 9 different language models on the SQuAD2.0 dataset, showing two key metrics: ANLS* (a document understanding metric) and Exact Match scores. The table reveals that LLama 3.1 405B achieves the highest ANLS* score of 80.37, followed closely by Claude 4 Sonnet at 79.65 and Arctic-Extract GA at 78.94. For Exact Match, Arctic-Extract GA leads with 76.57, narrowly ahead of LLama 3.1 405B at 76.56. Arctic-Extract achieves competitive results with much larger models while being significantly more resource-efficient, demonstrating strong performance relative to its size constraints.

\begin{table}[h]
\centering
\caption{Performance comparison on SQuAD2.0 dataset.}
\label{tab:model_comparison}
\begin{adjustbox}{center}
\begin{tabular}{@{}lcc@{}}
\toprule
\textbf{Model} & \textbf{ANLS*} & \textbf{Exact Match} \\ 
\midrule
\textbf{Arctic-Extract} & 78.94 & \textbf{76.57} \\
LLama 3.1 405B & \textbf{80.37} & 76.56 \\
Claude 4 Sonnet & 79.65 & 76.26 \\
LLama 4 Scout & 74.30 & 70.70 \\
GPT 5 & 74.27 & 70.27 \\
Qwen 2.5-VL & 66.76 & 63.47 \\
LLama 3.1 8B & 59.13 & 54.48 \\
Pixtral 12B & 55.23 & 52.01 \\
Mistral 7B & 13.50 & 10.42 \\
\bottomrule
\end{tabular}
\end{adjustbox}
\end{table}

\subsection{Visual datasets evaluation}

The performance comparison on internal (visual) datasets demonstrates Arctic-Extract's superior capability across diverse document understanding tasks. As shown in Table \ref{tab:performance_comparison}, Arctic-Extract achieves the highest average performance with a score of 64.2, significantly outperforming all other evaluated models. The model shows particularly strong performance on the CUAD dataset (70.6), Signatures dataset (53.0), and Checkboxes dataset (74.1), where it leads by substantial margins.

It is important to note that to ensure fair comparison across all models, approximately 20\% of datapoints were excluded from these results. These exclusions were necessary when certain models failed to process inputs due to size constraints or other limitations. The complete results, including cases where models received scores of 0 for failed processing, are presented in the appendix for transparency.

GPT5 demonstrates competitive performance on several datasets, particularly excelling in \textit{Real life cases} (88.8), GHEGA-based (54.2), and Infographics-based (74.3) datasets. Claude 4 Sonnet shows consistent but generally lower performance across most datasets, while Pixtral 12B and Qwen2.5-VL exhibit more variable results depending on the specific task requirements.

\subsection{Multilingual evaluation}

The multilingual evaluation demonstrates Arctic-Extract's superior performance across diverse languages and datasets. As shown in Table \ref{tab:multilingual_evaluation} (see Appendix), Arctic-Extract achieves the highest average score of 70.7 across all multilingual benchmarks, significantly outperforming other models including LLama 3.1 405B (67.0), GPT5 (66.8), and Claude 4 Sonnet (65.1).

Arctic-Extract particularly excels on European languages, achieving outstanding results on frenchQA-based dataset (90.5), GermanQuAD-based dataset (84.1), and SQuAD-it-based dataset (72.1). The model also demonstrates strong performance on Asian languages, leading in KorQuAD (79.8) and showing competitive results on JaQuAD (81.0).

On the MLQA benchmark, Arctic-Extract achieves first place in 4 out of 7 languages tested, including English (76.9), Spanish (65.6), and Chinese (60.7). For the xQuAD benchmark, the model dominates with first place performance in 9 out of 12 languages, including English (81.3), Greek (73.9), Spanish (74.8), and Romanian (74.8).

\subsection{Table extraction evaluation}

\begin{table}[H]
\centering
\caption{Performance comparison on table extraction (see Appendix \ref{business_table_example}) - excluding uncommon datapoints (only the ones that all the models were able to process are included).}
\label{tab:business_tables_results}
\begin{adjustbox}{center}
\begin{tabular}{@{}lc@{}}
\toprule
\textbf{Model} & \textbf{Business Tables dataset} \\ 
\midrule
Arctic-TILT & \textbf{0.725} $\pm$ 0.03 \\
Arctic-Extract & 0.720 $\pm$ 0.03 \\
Claude 4 Sonnet & 0.707 $\pm$ 0.03 \\
GPT5 & 0.651 $\pm$ 0.04 \\
Qwen2.5-VL 7B & 0.560 $\pm$ 0.03 \\
Pixtral 12B & 0.160 $\pm$ 0.02 \\
\bottomrule
\end{tabular}
\end{adjustbox}
\end{table}

\begin{table*}[htbp]
\centering
\small
\caption{Comparison of IDP offerings on DocVQA \cite{mathew2021docvqadatasetvqadocument}. "Natural Language Questions" indicates whether the offering can process questions in natural language directly on the document image, or if it requires text extraction as an intermediate step.}
\label{tab:idp_offerings_comparison}
\begin{adjustbox}{center}
\renewcommand{\arraystretch}{1.5} % Adjust row spacing
\begin{tabular}{@{}p{5cm}p{1.5cm}p{5cm}p{1cm}p{1cm}@{}}
\toprule
\textbf{IDP Offering} & \textbf{Natural Language Questions} & \textbf{Extraction Method} & \textbf{DocVQA} & \textbf{$\Delta$} \\ 
\midrule
Human Evaluation & N/A & N/A & 0.9811 & 0.0341 \\
\textbf{Snowflake Arctic-Extract} & \cmark & \textbf{Document image \& question.} & \textbf{0.947} & \textbf{0} \\
Azure OpenAI GPT-o3 & \cmark & Document image \& question. & 0.9339 & -0.0131 \\
GCP Gemini 2.5-Pro & \cmark & Document image \& question. & 0.9316 & -0.0154 \\
GCP Anthropic Claude 4 Sonnet & \cmark & Document image \& question. & 0.9119 & -0.0351 \\
Azure Document Intelligence + GPT-o3 & \xmark & Using Azure Document Intelligence to get the text, then asking GPT-o3. & 0.8853 & -0.0617 \\
GCP Document AI + Gemini & \xmark & Using Google Document AI to get the text, then asking Gemini. & 0.8497 & -0.0973 \\
AWS Textract & \cmark & Document image \& question. & 0.8313 & -0.1157 \\
\bottomrule
\end{tabular}
\end{adjustbox}
\end{table*}

The Table Extraction evaluation demonstrates Arctic-Extract's competitive performance in this specialized task. As shown in Table \ref{tab:business_tables_results}, Arctic-TILT slightly outperforms Arctic-Extract with a score of 0.725 compared to 0.720, representing a minimal difference within the confidence intervals. Both models significantly outperform other evaluated systems, with Claude 4 Sonnet achieving 0.707, GPT5 at 0.651, Qwen2.5-VL at 0.560, and Pixtral 12B at 0.160. It should be noted that approximately 50\% of the datapoints were excluded from this comparison to ensure fairness across all models, as some models failed to process certain inputs due to size constraints. The complete results with error=0 scoring for failed cases are available in the appendix. This performance reflects Arctic-Extract's strong capability in transforming unstructured document information into structured tabular formats, a complex task that requires understanding of spatial relationships, contextual information, and data organization patterns within business documents.

\subsection{DocVQA evaluation and comparison with other IDP offerings}

The table in this section (Table \ref{tab:idp_offerings_comparison}) compares various Intelligent Document Processing (IDP) offerings on the DocVQA \cite{mathew2021docvqadatasetvqadocument} benchmark. It evaluates their performance based on the DocVQA score and the difference relative to the Snowflake Arctic-Extract model. The table highlights that Snowflake Arctic-Extract achieves the highest score among automated systems (\textbf{0.947}), with human evaluation serving as the upper bound (\textbf{0.9811}). Other offerings, such as Azure OpenAI GPT-o3 and GCP Gemini 2.5-Pro, follow closely but with slightly lower scores. The table also distinguishes between systems that natively process document images and those that rely on intermediate text extraction, with the latter generally performing worse. This comparison underscores Arctic-Extract's strong performance in document understanding tasks.

\section{Summary}

In this paper, we introduced Arctic-Extract, a state-of-the-art model designed for advanced document understanding tasks. Arctic-Extract is built upon the Qwen2.5-VL architecture, which features a unique token compression mechanism that enables efficient processing of large context windows. This capability allows the model to handle extensive documents, such as those spanning multiple pages, without compromising accuracy. The model was fine-tuned using LoRA, a low-rank adaptation technique, and subsequently merged and quantized to 4-bit precision. This optimization ensures that Arctic-Extract can be deployed on widely available hardware while maintaining high performance and efficiency. Notably, the model's compact size, requiring only 6.6 GiB of memory, makes it highly suitable for deployment on devices with limited resources.

Arctic-Extract was rigorously evaluated across a wide range of benchmarks, including SQuAD2.0, DocVQA, and various internal datasets. These evaluations demonstrated its exceptional capabilities in tasks such as question answering, table extraction, and multilingual document understanding. The model consistently delivered strong results, outperforming several existing state-of-the-art models, including various versions of GPT5, Qwen2.5-VL 7B, Pixtral 12B, and Claude 4 Sonnet. Arctic-Extract's ability to process diverse document types and languages highlights its versatility and adaptability to real-world scenarios.

A key strength of Arctic-Extract lies in its multilingual capabilities, as evidenced by its performance on datasets spanning numerous languages. This makes it particularly well-suited for applications requiring global reach and support for diverse linguistic contexts. Additionally, the model's robust performance in table extraction tasks underscores its utility in processing structured data, a critical requirement in many business and financial applications.

The results of this study underscore Arctic-Extract's robustness, scalability, compact size, and versatility, making it a compelling solution for Intelligent Document Processing (IDP) applications. By combining cutting-edge architectural innovations with practical deployment considerations, Arctic-Extract sets a new benchmark for document understanding models and paves the way for further advancements in the field.
% Bibliography entries for the entire Anthology, followed by custom entries
%\bibliography{anthology,custom}
% Custom bibliography entries only
\bibliography{bibliography}

@inproceedings{10.1007/978-3-030-03840-3_29,
  author    = {Croce, Danilo and Zelenanska, Alexandra and Basili, Roberto},
  editor    = {Ghidini, Chiara and Magnini, Bernardo and Passerini, Andrea and Traverso, Paolo},
  title     = {Neural Learning for Question Answering in {Italian}},
  booktitle = {AI*IA 2018 -- Advances in Artificial Intelligence},
  year      = {2018},
  publisher = {Springer International Publishing},
  address   = {Cham},
  pages     = {389--402},
  isbn      = {978-3-030-03840-3}
}

@article{Zanibbi2004,
  author={Zanibbi, Richard and Blostein, Dorothea and Cordy, James R.},
  title={A survey of table recognition},
  journal={Document Analysis and Recognition},
  year      = {2004},
  month     = {Mar},
  day       = {01},
  volume    = {7},
  number    = {1},
  pages     = {1-16},
  issn      = {1433-2825},
  doi       = {10.1007/s10032-004-0120-9},
  url       = {https://doi.org/10.1007/s10032-004-0120-9}
}

@inproceedings{10.1145/3587259.3627548,
  author    = {Tuora, Ryszard and Zwierzchowska, Aleksandra and Zawadzka-Paluektau, Natalia and Klamra, Cezary and Kobyli\'{n}ski, \L{}ukasz},
  title     = {{PoQuAD} - The {Polish} Question Answering Dataset - Description and Analysis},
  year      = {2023},
  isbn      = {9798400701412},
  publisher = {Association for Computing Machinery},
  address   = {New York, NY, USA},
  url       = {https://doi.org/10.1145/3587259.3627548},
  doi       = {10.1145/3587259.3627548},
  booktitle = {Proceedings of the 12th Knowledge Capture Conference 2023},
  pages     = {105–113},
  numpages  = {9},
  location  = {Pensacola, FL, USA},
  series    = {K-CAP '23}
}

@inproceedings{2019TabFactA,
  title     = {{TabFact} : A Large-scale Dataset for Table-based Fact Verification},
  author    = {Wenhu Chen and Hongmin Wang and Jianshu Chen and Yunkai Zhang and Hong Wang and Shiyang Li and Xiyou Zhou and William Yang Wang},
  booktitle = {International Conference on Learning Representations (ICLR)},
  address   = {Addis Ababa, Ethiopia},
  month     = {April},
  year      = {2020}
}

@misc{agrawal2024pixtral12b,
  title         = {Pixtral 12B},
  author        = {Pravesh Agrawal and Szymon Antoniak and Emma Bou Hanna and Baptiste Bout and Devendra Chaplot and Jessica Chudnovsky and Diogo Costa and Baudouin De Monicault and Saurabh Garg and Theophile Gervet and Soham Ghosh and Amélie Héliou and Paul Jacob and Albert Q. Jiang and Kartik Khandelwal and Timothée Lacroix and Guillaume Lample and Diego Las Casas and Thibaut Lavril and Teven Le Scao and Andy Lo and William Marshall and Louis Martin and Arthur Mensch and Pavankumar Muddireddy and Valera Nemychnikova and Marie Pellat and Patrick Von Platen and Nikhil Raghuraman and Baptiste Rozière and Alexandre Sablayrolles and Lucile Saulnier and Romain Sauvestre and Wendy Shang and Roman Soletskyi and Lawrence Stewart and Pierre Stock and Joachim Studnia and Sandeep Subramanian and Sagar Vaze and Thomas Wang and Sophia Yang},
  year          = {2024},
  eprint        = {2410.07073},
  archiveprefix = {arXiv},
  primaryclass  = {cs.CV},
  url           = {https://arxiv.org/abs/2410.07073}
}

@misc{bai2023qwenvlversatilevisionlanguagemodel,
  title         = {{Qwen-VL}: A Versatile Vision-Language Model for Understanding, Localization, Text Reading, and Beyond},
  author        = {Jinze Bai and Shuai Bai and Shusheng Yang and Shijie Wang and Sinan Tan and Peng Wang and Junyang Lin and Chang Zhou and Jingren Zhou},
  year          = {2023},
  eprint        = {2308.12966},
  archiveprefix = {arXiv},
  primaryclass  = {cs.CV},
  url           = {https://arxiv.org/abs/2308.12966}
}

@misc{bai2025qwen25vltechnicalreport,
  title         = {{Qwen2.5-VL} Technical Report},
  author        = {Shuai Bai and Keqin Chen and Xuejing Liu and Jialin Wang and Wenbin Ge and Sibo Song and Kai Dang and Peng Wang and Shijie Wang and Jun Tang and Humen Zhong and Yuanzhi Zhu and Mingkun Yang and Zhaohai Li and Jianqiang Wan and Pengfei Wang and Wei Ding and Zheren Fu and Yiheng Xu and Jiabo Ye and Xi Zhang and Tianbao Xie and Zesen Cheng and Hang Zhang and Zhibo Yang and Haiyang Xu and Junyang Lin},
  year          = {2025},
  eprint        = {2502.13923},
  archiveprefix = {arXiv},
  primaryclass  = {cs.CV},
  url           = {https://arxiv.org/abs/2502.13923}
}

@inproceedings{borchmann2021due,
  title     = {{DUE}: End-to-End Document Understanding Benchmark},
  author    = {{\L}ukasz Borchmann and Micha{\l} Pietruszka and Tomasz Stanislawek and Dawid Jurkiewicz and Micha{\l} Turski and Karolina Szyndler and Filip Grali{\'n}ski},
  booktitle = {Thirty-fifth Conference on Neural Information Processing Systems Datasets and Benchmarks Track (Round 2)},
  year      = {2021},
  url       = {https://openreview.net/forum?id=rNs2FvJGDK}
}

@misc{borchmann2024arctictiltbusinessdocumentunderstanding,
  title         = {{Arctic-TILT}. Business Document Understanding at Sub-Billion Scale},
  author        = {{\L}ukasz Borchmann and Michał Pietruszka and Wojciech Jaśkowski and Dawid Jurkiewicz and Piotr Halama and Paweł Józiak and Łukasz Garncarek and Paweł Liskowski and Karolina Szyndler and Andrzej Gretkowski and Julita Ołtusek and Gabriela Nowakowska and Artur Zawłocki and Łukasz Duhr and Paweł Dyda and Michał Turski},
  year          = {2024},
  eprint        = {2408.04632},
  archiveprefix = {arXiv},
  primaryclass  = {cs.CL},
  url           = {https://arxiv.org/abs/2408.04632}
}

@misc{carrino2019automatic,
  title         = {Automatic {Spanish} Translation of the {SQuAD} Dataset for Multilingual Question Answering},
  author        = {Casimiro Pio Carrino and Marta R. Costa-jussà and José A. R. Fonollosa},
  year          = {2019},
  eprint        = {1912.05200},
  archiveprefix = {arXiv},
  primaryclass  = {cs.CL}
}

@dataset{cruz_2022_6371710,
  author    = {Cruz, Francisco and Castelli, Mauro},
  title     = {Dataset of invoices and receipts including annotation of relevant fields},
  month     = mar,
  year      = 2022,
  publisher = {Zenodo},
  doi       = {10.5281/zenodo.6371710},
  url       = {https://doi.org/10.5281/zenodo.6371710}
}

@misc{dosovitskiy2021imageworth16x16words,
  title         = {An Image is Worth 16x16 Words: Transformers for Image Recognition at Scale},
  author        = {Alexey Dosovitskiy and Lucas Beyer and Alexander Kolesnikov and Dirk Weissenborn and Xiaohua Zhai and Thomas Unterthiner and Mostafa Dehghani and Matthias Minderer and Georg Heigold and Sylvain Gelly and Jakob Uszkoreit and Neil Houlsby},
  year          = {2021},
  eprint        = {2010.11929},
  archiveprefix = {arXiv},
  primaryclass  = {cs.CV},
  url           = {https://arxiv.org/abs/2010.11929}
}

@misc{frenchQA2023,
  author       = { {ALBAR, Boris and BEDU, Pierre and BOURDOIS, Loïck} },
  organization = { {Centre Aquitain des Technologies de l'Information et Electroniques} },
  title        = {{frenchQA} (Revision 6249cd5)},
  year         = 2023,
  url          = { https://huggingface.co/CATIE-AQ/frenchQA },
  doi          = { 10.57967/hf/0862 },
  publisher    = { Hugging Face }
}

@misc{githubGitHubCasperhansenAutoAWQ,
  author       = {casper-hansen},
  title        = {{G}it{H}ub - casper-hansen/{A}uto{A}{W}{Q}: {A}uto{A}{W}{Q} implements the {A}{W}{Q} algorithm for 4-bit quantization with a 2x speedup during inference. {D}ocumentation: --- github.com},
  howpublished = {\url{https://github.com/casper-hansen/AutoAWQ}},
  year         = {},
  note         = {[Accessed 07-10-2025]}
}

@misc{githubGitHubPiEspositobrquad20,
  author       = {Pi Esposito},
  title        = {{G}it{H}ub - pi{E}sposito/br-quad-2.0: {S}tanford {Q}uestion {A}nswering {D}ataset ({S}{Q}u{A}{D}) 2.0 translated to {B}razilian {P}ortuguese ({P}{T}-{B}{R}) language. --- github.com},
  howpublished = {\url{https://github.com/piEsposito/br-quad-2.0}},
  year         = {},
  note         = {[Accessed 06-10-2025]}
}

@misc{hu2021loralowrankadaptationlarge,
  title         = {{LoRA}: Low-Rank Adaptation of Large Language Models},
  author        = {Edward J. Hu and Yelong Shen and Phillip Wallis and Zeyuan Allen-Zhu and Yuanzhi Li and Shean Wang and Lu Wang and Weizhu Chen},
  year          = {2021},
  eprint        = {2106.09685},
  archiveprefix = {arXiv},
  primaryclass  = {cs.CL},
  url           = {https://arxiv.org/abs/2106.09685}
}

@inproceedings{kwon2023efficient,
  title     = {Efficient Memory Management for Large Language Model Serving with {PagedAttention}},
  author    = {Woosuk Kwon and Zhuohan Li and Siyuan Zhuang and Ying Sheng and Lianmin Zheng and Cody Hao Yu and Joseph E. Gonzalez and Hao Zhang and Ion Stoica},
  booktitle = {Proceedings of the ACM SIGOPS 29th Symposium on Operating Systems Principles},
  year      = {2023}
}

@misc{lin2024awqactivationawareweightquantization,
  title         = {{AWQ}: Activation-aware Weight Quantization for {LLM} Compression and Acceleration},
  author        = {Ji Lin and Jiaming Tang and Haotian Tang and Shang Yang and Wei-Ming Chen and Wei-Chen Wang and Guangxuan Xiao and Xingyu Dang and Chuang Gan and Song Han},
  year          = {2024},
  eprint        = {2306.00978},
  archiveprefix = {arXiv},
  primaryclass  = {cs.CL},
  url           = {https://arxiv.org/abs/2306.00978}
}

@misc{liu2023visualinstructiontuning,
  title         = {Visual Instruction Tuning},
  author        = {Haotian Liu and Chunyuan Li and Qingyang Wu and Yong Jae Lee},
  year          = {2023},
  eprint        = {2304.08485},
  archiveprefix = {arXiv},
  primaryclass  = {cs.CV},
  url           = {https://arxiv.org/abs/2304.08485}
}

@misc{loshchilov2019decoupledweightdecayregularization,
  title         = {Decoupled Weight Decay Regularization},
  author        = {Ilya Loshchilov and Frank Hutter},
  year          = {2019},
  eprint        = {1711.05101},
  archiveprefix = {arXiv},
  primaryclass  = {cs.LG},
  url           = {https://arxiv.org/abs/1711.05101}
}

@article{maria,
  author  = {Asier Gutiérrez-Fandiño and Jordi Armengol-Estapé and Marc Pàmies and Joan Llop-Palao and Joaquin Silveira-Ocampo and Casimiro Pio Carrino and Carme Armentano-Oller and Carlos Rodriguez-Penagos and Aitor Gonzalez-Agirre and Marta Villegas},
  title   = {MarIA: {Spanish} Language Models},
  journal = {Procesamiento del Lenguaje Natural},
  volume  = {68},
  number  = {0},
  year    = {2022},
  issn    = {1989-7553},
  url     = {http://journal.sepln.org/sepln/ojs/ojs/index.php/pln/article/view/6405},
  pages   = {39--60}
}

@misc{mathew2021docvqadatasetvqadocument,
  title         = {{DocVQA}: A Dataset for {VQA} on Document Images},
  author        = {Minesh Mathew and Dimosthenis Karatzas and C. V. Jawahar},
  year          = {2021},
  eprint        = {2007.00398},
  archiveprefix = {arXiv},
  primaryclass  = {cs.CV},
  url           = {https://arxiv.org/abs/2007.00398}
}

@misc{mathew2021infographicvqa,
  title         = {{InfographicVQA}},
  author        = {Minesh Mathew and Viraj Bagal and Rubèn Pérez Tito and Dimosthenis Karatzas and Ernest Valveny and C. V Jawahar},
  year          = {2021},
  eprint        = {2104.12756},
  archiveprefix = {arXiv},
  primaryclass  = {cs.CV},
  url           = {https://arxiv.org/abs/2104.12756}
}

@misc{möller2021germanquad,
  title         = {{GermanQuAD} and {GermanDPR}: Improving Non-{English} Question Answering and Passage Retrieval},
  author        = {Timo Möller and Julian Risch and Malte Pietsch},
  year          = {2021},
  eprint        = {2104.12741},
  archiveprefix = {arXiv},
  primaryclass  = {cs.CL}
}

@misc{peer2025anlsuniversaldocument,
  title         = {{ANLS*} -- A Universal Document Processing Metric for Generative Large Language Models},
  author        = {David Peer and Philemon Schöpf and Volckmar Nebendahl and Alexander Rietzler and Sebastian Stabinger},
  year          = {2025},
  eprint        = {2402.03848},
  archiveprefix = {arXiv},
  primaryclass  = {cs.CL},
  url           = {https://arxiv.org/abs/2402.03848}
}

@article{sanchez2022idsem,
  title     = {{IDSEM}, an invoices database of the {Spanish} electricity market},
  author    = {S{\'a}nchez, Javier and Salgado, Agust{\'\i}n and Garc{\'\i}a, Alejandro and Monz{\'o}n, Nelson},
  journal   = {Scientific data},
  volume    = {9},
  number    = {1},
  pages     = {786},
  year      = {2022},
  publisher = {Nature Publishing Group UK London}
}

@inbook{Stanis_awek_2021,
  title     = {Kleister: Key Information Extraction Datasets Involving Long Documents with Complex Layouts},
  isbn      = {9783030865498},
  issn      = {1611-3349},
  url       = {http://dx.doi.org/10.1007/978-3-030-86549-8_36},
  doi       = {10.1007/978-3-030-86549-8_36},
  booktitle = {Document Analysis and Recognition – ICDAR 2021},
  publisher = {Springer International Publishing},
  author    = {Stanisławek, Tomasz and Graliński, Filip and Wróblewska, Anna and Lipiński, Dawid and Kaliska, Agnieszka and Rosalska, Paulina and Topolski, Bartosz and Biecek, Przemysław},
  year      = {2021},
  pages     = {564–579}
}

@misc{vanlandeghem2023documentunderstandingdatasetevaluation,
  title         = {Document Understanding Dataset and Evaluation ({DUDE})},
  author        = {Jordy Van Landeghem and Rubén Tito and Łukasz Borchmann and Michał Pietruszka and Paweł Józiak and Rafał Powalski and Dawid Jurkiewicz and Mickaël Coustaty and Bertrand Ackaert and Ernest Valveny and Matthew Blaschko and Sien Moens and Tomasz Stanisławek},
  year          = {2023},
  eprint        = {2305.08455},
  archiveprefix = {arXiv},
  primaryclass  = {cs.CV},
  url           = {https://arxiv.org/abs/2305.08455}
}

@misc{wu2024deepseekvl2mixtureofexpertsvisionlanguagemodels,
  title         = {{DeepSeek-VL2}: Mixture-of-Experts Vision-Language Models for Advanced Multimodal Understanding},
  author        = {Zhiyu Wu and Xiaokang Chen and Zizheng Pan and Xingchao Liu and Wen Liu and Damai Dai and Huazuo Gao and Yiyang Ma and Chengyue Wu and Bingxuan Wang and Zhenda Xie and Yu Wu and Kai Hu and Jiawei Wang and Yaofeng Sun and Yukun Li and Yishi Piao and Kang Guan and Aixin Liu and Xin Xie and Yuxiang You and Kai Dong and Xingkai Yu and Haowei Zhang and Liang Zhao and Yisong Wang and Chong Ruan},
  year          = {2024},
  eprint        = {2412.10302},
  archiveprefix = {arXiv},
  primaryclass  = {cs.CV},
  url           = {https://arxiv.org/abs/2412.10302}
}

@misc{yuan2021florencenewfoundationmodel,
  title         = {Florence: A New Foundation Model for Computer Vision},
  author        = {Lu Yuan and Dongdong Chen and Yi-Ling Chen and Noel Codella and Xiyang Dai and Jianfeng Gao and Houdong Hu and Xuedong Huang and Boxin Li and Chunyuan Li and Ce Liu and Mengchen Liu and Zicheng Liu and Yumao Lu and Yu Shi and Lijuan Wang and Jianfeng Wang and Bin Xiao and Zhen Xiao and Jianwei Yang and Michael Zeng and Luowei Zhou and Pengchuan Zhang},
  year          = {2021},
  eprint        = {2111.11432},
  archiveprefix = {arXiv},
  primaryclass  = {cs.CV},
  url           = {https://arxiv.org/abs/2111.11432}
}

@misc{zhao2024swiftascalablelightweightinfrastructure,
  title         = {{SWIFT}: A Scalable lightWeight Infrastructure for Fine-Tuning},
  author        = {Yuze Zhao and Jintao Huang and Jinghan Hu and Xingjun Wang and Yunlin Mao and Daoze Zhang and Zeyinzi Jiang and Zhikai Wu and Baole Ai and Ang Wang and Wenmeng Zhou and Yingda Chen},
  year          = {2024},
  eprint        = {2408.05517},
  archiveprefix = {arXiv},
  primaryclass  = {cs.CL},
  url           = {https://arxiv.org/abs/2408.05517}
}

@inproceedings{Zhu_2022,
  series     = {MM ’22},
  title      = {Towards Complex Document Understanding By Discrete Reasoning},
  url        = {http://dx.doi.org/10.1145/3503161.3548422},
  doi        = {10.1145/3503161.3548422},
  booktitle  = {Proceedings of the 30th ACM International Conference on Multimedia},
  publisher  = {ACM},
  author     = {Zhu, Fengbin and Lei, Wenqiang and Feng, Fuli and Wang, Chao and Zhang, Haozhou and Chua, Tat-Seng},
  year       = {2022},
  month      = oct,
  pages      = {4857–4866},
  collection = {MM ’22}
}

@misc{kim2022ocrfreedocumentunderstandingtransformer,
      title={{OCR-free} Document Understanding Transformer}, 
      author={Geewook Kim and Teakgyu Hong and Moonbin Yim and Jeongyeon Nam and Jinyoung Park and Jinyeong Yim and Wonseok Hwang and Sangdoo Yun and Dongyoon Han and Seunghyun Park},
      year={2022},
      eprint={2111.15664},
      archivePrefix={arXiv},
      primaryClass={cs.LG},
      url={https://arxiv.org/abs/2111.15664}, 
}

\onecolumn
\clearpage
\appendix
\section{Internal Datasets benchmarked in the study}

\paragraph{Datasets Overview.}
In this study, we have evaluated the model on 10 internal datasets, that are not publicaly available. These datasets cover a variety of document types and tasks, including form understanding, invoice processing, and contract analysis. Each dataset consists of scanned documents or PDFs, along with corresponding annotations for tasks such as entity extraction, classification, and question answering. In this section, each of the datasets will be described. The quantitive details of the datasets are summarized in Table \ref{tab:internal_datasets}.
\begin{table}[H]
\centering
\small
\caption{Quantitative details of the internal datasets benchmarked in the study.}
\label{tab:internal_datasets}
\begin{adjustbox}{center}
\begin{tabular}{@{}lccccccccc@{}}
\toprule
\multicolumn{1}{c}{\textbf{Dataset}} & \multicolumn{1}{c}{\textbf{Documents}} & \multicolumn{4}{c}{\textbf{Pages}} & \multicolumn{4}{c}{\textbf{Questions}} \\ 
\cmidrule(lr){3-6} \cmidrule(lr){7-10}
& & Min & Max & Mean ± std & Total & Min & Max & Mean ± std & Total \\ 
\midrule
CUAD-based dataset & 169 & 1 & 115 & 17.7 ± 20.1 & 2,983 & 35 & 41 & 38.6 ± 1.5 & 6,517 \\
Real life cases dataset & 111 & 1 & 124 & 18.8 ± 37.8 & 2,092 & 3 & 32 & 14.6 ± 9.7 & 1,616 \\
Hard bussiness cases dataset & 12 & 1 & 9 & 3.5 ± 3.5 & 42 & 1 & 11 & 3.8 ± 4.1 & 45 \\
Agreements dataset & 49 & 3 & 75 & 20.2 ± 15.8 & 992 & 15 & 24 & 17.9 ± 2.1 & 879 \\
GHEGA-based dataset & 42 & 1 & 1 & 1.0 ± 0.0 & 42 & 2 & 11 & 9.0 ± 2.3 & 379 \\
InfographicsVQA-based dataset & 579 & 1 & 1 & 1.0 ± 0.0 & 579 & 1 & 29 & 5.7 ± 4.6 & 3,288 \\
Financial dataset & 39 & 1 & 2 & 1.0 ± 0.2 & 40 & 6 & 13 & 11.5 ± 2.3 & 448 \\
SEC-based dataset & 48 & 23 & 121 & 72.1 ± 30.5 & 3,462 & 4 & 10 & 6.8 ± 2.4 & 328 \\
Signatures dataset & 88 & 1 & 54 & 10.6 ± 13.6 & 931 & 5 & 5 & 5.0 ± 0.0 & 440 \\
Checkboxes dataset & 36 & 1 & 102 & 11.0 ± 18.8 & 395 & 1 & 5 & 3.4 ± 1.7 & 121 \\
\bottomrule
\end{tabular}
\end{adjustbox}
\end{table}
\paragraph{CUAD-based dataset.}
The CUAD-based dataset, sourced from the EDGAR system used by the U.S. Securities and Exchange Commission (SEC), comprises contracts designed to benchmark contract understanding tasks. These documents are rich in legal language and structure, making them ideal for evaluating models on contract analysis.

\paragraph{Real life cases dataset.}
This dataset includes a variety of semi-structured documents such as forms, invoices, receipts, and reports. It reflects real-world Document Understanding use cases, offering diverse layouts and content types for robust model evaluation.

\paragraph{Hard bussiness cases dataset.}
Extracted and separated from the \textit{Real life cases dataset}, this collection focuses on documents that are significantly harder to interpret. These challenges arise due to factors like poor scan quality, hard-to-read handwriting, or other obstacles that test the limits of Document Understanding systems.

\paragraph{Agreements dataset.}
This dataset consists of various types of agreements, providing a focused benchmark for models tasked with understanding and extracting information from legal and contractual documents.

\paragraph{GHEGA-based dataset.}
Derived from the GHEGA dataset, this collection includes patents and is tailored for tasks involving technical and structured document analysis. It provides a unique challenge due to the dense and specialized content of patent documents.

\paragraph{Infographics-based dataset.}
Based on the InfographicVQA dataset from the DocVQA challenge, this dataset includes visually rich infographics. It is designed to test models on tasks requiring the integration of visual and textual information.

\paragraph{Financial dataset.}
This dataset contains both real and generated financial documents, including receipts, invoices, and earning statements. The fictitious data ensures privacy while providing realistic scenarios for financial document processing.

\paragraph{SEC-based dataset.}
Comprising publicly available documents from the United States Securities and Exchange Commission, this dataset is ideal for evaluating models on financial and regulatory document analysis.

\paragraph{Signatures dataset.}
This dataset includes documents that feature signatures or designated signature areas. It is particularly useful for tasks involving signature detection and verification.

\paragraph{Checkboxes dataset.}
This dataset contains diverse documents annotated with Q\&A pairs about checkboxes. It is designed to benchmark models on fine-grained tasks such as detecting checkbox states and associating them with nearby textual descriptions.

\section{Business Tables dataset}
\label{business_table_example}

The Business Tables dataset is another internal dataset, however, as its primary focus is not extracting single values, or lists, we decided to have it as separate entity. It consists of various business-related documents, including invoices, purchase orders, and financial reports, all structured in tabular formats.

\begin{table}[H]
\centering
\caption{Quantitative details of the Business Tables dataset benchmarked in the study.}
\begin{adjustbox}{center}
\begin{tabular}{@{}cccccccccc@{}}
\toprule
\textbf{Documents} & \multicolumn{4}{c}{\textbf{Pages}} & \multicolumn{4}{c}{\textbf{Questions}} \\ 
\cmidrule(lr){2-5} \cmidrule(lr){6-9}
& Min & Max & Mean ± std & Total & Min & Max & Mean ± std & Total \\ 
\midrule
282 & 1 & 60 & 13.6 ± 16.8 & 3,832 & 1 & 35 & 3.1 ± 4.0 & 878 \\
\bottomrule
\end{tabular}
\end{adjustbox}
\end{table}

An example of the table would be extracting table with the following header:

\begin{displayquote}
\footnotesize
\texttt{Sale Month | Sale Year | Price | OR Book\/Page | Sale Type | Owner}
\end{displayquote}

And the expected model output would be:

\begin{displayquote}
\footnotesize
\texttt{JAN|2021|\$2,175,000|None|WARRANTY DEED|None \textbackslash n\\ JUN|2015|\$1,682,500|None|QUIT CLAIM|None \textbackslash n\\ DEC|2002|\$125,000|None|WARRANTY DEED|None \textbackslash n\\ DEC|1988|\$100|None|WARRANTY DEED|Andrew \textbackslash n\\ FEB|1987|\$128,600|None|WARRANTY DEED|None \textbackslash n\\ SEP|1986|\$2,200,000|None|WARRANTY DEED|None \textbackslash n\\ JAN|1985|\$1,670,000|None|WARRANTY DEED|None \textbackslash n\\ JUN|1981|\$2,120,000|None|WARRANTY DEED|None \textbackslash n\\ FEB|1971|\$1,250,000|None|None|None}
\end{displayquote}

That would be rendered to the following table:

\begin{table}[H]
\centering
\caption{Extracted Business Table Example.}
\begin{adjustbox}{center}
  \small % Set smaller font size
\begin{tabular}{|l|l|l|l|l|l|}
\hline
\textbf{Sale Month} & \textbf{Sale Year} & \textbf{Price} & \textbf{OR Book/Page} & \textbf{Sale Type} & \textbf{Owner} \\ \hline
JAN & 2021 & \$2,175,000 &  & WARRANTY DEED &  \\ \hline
JUN & 2015 & \$1,682,500 &  & QUIT CLAIM &  \\ \hline
DEC & 2002 & \$125,000 &  & WARRANTY DEED &  \\ \hline
DEC & 1988 & \$100 &  & WARRANTY DEED & Andrew \\ \hline
FEB & 1987 & \$128,600 &  & WARRANTY DEED &  \\ \hline
SEP & 1986 & \$2,200,000 &  & WARRANTY DEED &  \\ \hline
JAN & 1985 & \$1,670,000 &  & WARRANTY DEED &  \\ \hline
JUN & 1981 & \$2,120,000 &  & WARRANTY DEED &  \\ \hline
FEB & 1971 & \$1,250,000 &  &  &  \\ \hline
\end{tabular}
\end{adjustbox}
\end{table}

\section{Experiments - full results}

\begin{table*}[htbp]
\centering
\caption{Performance comparison across internal datasets - excluding uncommon datapoints (only the ones that all the models were able to process are included).}
\label{tab:performance_comparison}
\begin{adjustbox}{center}
\renewcommand{\arraystretch}{1.2}
\begin{tabular}{@{}lcccccc@{}}
\toprule
\textbf{Dataset} & \textbf{Arctic-Extract} & \textbf{Claude 4 Sonnet} & \textbf{GPT5} & \textbf{Pixtral 12B} & \textbf{Qwen 2.5-VL} & \textbf{TILT 1.10} \\
\midrule
CUAD & \textbf{70.6} $\pm$ 1.2 & 6.1 $\pm$ 0.6 & 15.4 $\pm$ 1.0 & 28.1 $\pm$ 1.3 & 10.3 $\pm$ 0.9 & 4.3 $\pm$ 0.6 \\
Real life cases & 84.4 $\pm$ 1.9 & 84.7 $\pm$ 2.0 & \textbf{88.8} $\pm$ 1.6 & 60.9 $\pm$ 2.5 & 80.6 $\pm$ 2.0 & 76.7 $\pm$ 2.2 \\
Hard bussiness cases & 69.1 $\pm$ 13.4 & 65.8 $\pm$ 13.0 & 67.9 $\pm$ 13.9 & 34.2 $\pm$ 11.3 & \textbf{69.6} $\pm$ 12.4 & 58.8 $\pm$ 13.1 \\
Agreements & 58.6 $\pm$ 3.9 & 54.6 $\pm$ 3.9 & \textbf{62.0} $\pm$ 3.8 & 34.1 $\pm$ 3.6 & 54.3 $\pm$ 3.8 & 44.6 $\pm$ 3.9 \\
GHEGA-based & 44.3 $\pm$ 4.8 & 45.1 $\pm$ 4.6 & \textbf{54.2} $\pm$ 4.7 & 32.2 $\pm$ 4.5 & 45.2 $\pm$ 4.4 & 30.7 $\pm$ 4.3 \\
Infographics-based & 73.0 $\pm$ 1.4 & 56.7 $\pm$ 1.7 & \textbf{74.3} $\pm$ 1.4 & 29.5 $\pm$ 1.4 & 67.3 $\pm$ 1.6 & 54.7 $\pm$ 1.6 \\
Financial & 50.8 $\pm$ 4.6 & 47.2 $\pm$ 4.3 & \textbf{52.6} $\pm$ 4.7 & 38.2 $\pm$ 4.6 & 46.3 $\pm$ 4.5 & 40.9 $\pm$ 4.3 \\
Signatures & \textbf{53.0} $\pm$ 4.8 & 36.0 $\pm$ 4.3 & 35.0 $\pm$ 4.4 & 30.5 $\pm$ 4.1 & 21.7 $\pm$ 3.9 & 30.8 $\pm$ 4.2 \\
Checkboxes & 74.1 $\pm$ 8.4 & 65.2 $\pm$ 9.1 & \textbf{76.0} $\pm$ 7.7 & 35.2 $\pm$ 8.2 & 62.3 $\pm$ 8.9 & 58.3 $\pm$ 9.3 \\
\midrule
\textbf{Average} & \textbf{64.2} & 51.3 & 58.5 & 35.9 & 50.8 & 44.4 \\
\bottomrule
\end{tabular}
\end{adjustbox}
\end{table*}

In our main study, we deliberately excluded certain datapoints to ensure a fair and balanced comparison across all evaluated models. These exclusions were necessary because some models encountered processing failures due to various constraints, including:

\begin{itemize}
  \item Input size limitations that prevented processing of large documents
  \item Memory constraints that caused processing errors
  \item API timeouts or processing errors for cloud-based models
  \item Lack of support for certain input formats or languages
\end{itemize}

To maintain methodological rigor, we excluded any datapoints that could not be processed by one or more models from the main comparison tables. This approach ensures that all reported scores are based on the same subset of data, enabling meaningful performance comparisons.

However, for completeness and transparency, the following tables present the full experimental results where models that failed to process a given datapoint received a score of 0 for that example. This scoring approach reflects the real-world deployment scenario where processing failures result in no extracted information. These comprehensive results demonstrate the practical advantages of Arctic-Extract's ability to handle large documents and challenging inputs that cause other models to fail entirely. Note that many of the failures observed for cloud-based models (GPT5 and Claude 4 Sonnet) are primarily attributed to CortexAI (that we used for the internal datasets, that had to stay within our security perimeter) platform limitations rather than inherent model capabilities.

The comprehensive Business Tables evaluation (Table \ref{tab:business_tables}) shows the performance when failed processing attempts are scored as 0, revealing the true deployment challenges faced by different models. Arctic-TILT leads with a score of 0.70, closely followed by Arctic-Extract at 0.68. The table demonstrates significant performance degradation for cloud-based models compared to the filtered results shown earlier.

Notably, Qwen2.5-VL maintains reasonable performance at 0.50, while Claude 4 Sonnet and GPT5 show substantial drops to 0.41 and 0.40 respectively. Pixtral 12B shows consistent performance at 0.16, indicating stable but limited capability for this task.

\begin{table}[H]
\centering
\caption{Performance comparison on table extraction task - error yielding score 0.}
\label{tab:business_tables}
\begin{adjustbox}{center}
\begin{tabular}{@{}lc@{}}
\toprule
\textbf{Model} & \textbf{Business Tables dataset} \\ 
\midrule
Arctic-TILT & \textbf{0.70} $\pm$ 0.02 \\
Arctic-Extract & 0.68 $\pm$ 0.02 \\
Qwen2.5-VL 7B & 0.50 $\pm$ 0.02 \\
Claude 4 Sonnet & 0.41 $\pm$ 0.03 \\
GPT5 & 0.40 $\pm$ 0.03 \\
Pixtral 12B & 0.16 $\pm$ 0.02 \\
\bottomrule
\end{tabular}
\end{adjustbox}
\end{table}

The table reveals that both GPT5 and Claude 4 Sonnet experience significant processing failures on the Business Tables dataset, with error rates of 42.79\% and 43.38\% respectively.

\begin{table}[H]
\centering
\caption{Error rates of GPT and Claude 4 Sonnet on Business Tables dataset.}
\begin{adjustbox}{center}
\begin{tabular}{@{}lc@{}}
\toprule
\textbf{Model} & \textbf{Error rate} \\ 
\midrule
GPT & 42.79\% \\
Claude 4 Sonnet & 43.38\% \\
\bottomrule
\end{tabular}
\end{adjustbox}
\end{table}

The comprehensive performance comparison (Table \ref{tab:performance_comparison_full}) demonstrates Arctic-Extract's robust capabilities when accounting for all processing attempts, including failures scored as 0. Arctic-Extract achieves the highest average performance at 66.5, significantly outperforming all other models. This comprehensive evaluation reveals the practical deployment advantages of Arctic-Extract's ability to reliably process challenging documents that cause other models to fail entirely.

Arctic-Extract shows exceptional performance on the SEC-based dataset (92.7), where both GPT5 and Claude 4 Sonnet completely failed to process any documents (0.0 scores). This stark difference highlights Arctic-Extract's superior capability in handling large, complex multi-page financial documents. The model also leads on several other datasets including CUAD (68.2), \textit{Real life cases} (84.1), and Agreements (57.4).

\begin{table*}[htbp]
\centering
\caption{Performance comparison across internal datasets - error yielding score 0.}
\label{tab:performance_comparison_full}
\begin{adjustbox}{center}
\renewcommand{\arraystretch}{1.2}
\begin{tabular}{@{}lcccccc@{}}
\toprule
\textbf{Dataset} & \textbf{Arctic-Extract} & \textbf{Claude 4 Sonnet} & \textbf{GPT5} & \textbf{Pixtral 12B} & \textbf{Qwen2.5-VL} & \textbf{TILT 1.10} \\
\midrule
CUAD & \textbf{68.2} $\pm$ 1.1 & 4.4 $\pm$ 0.5 & 15.2 $\pm$ 0.9 & 26.1 $\pm$ 1.0 & 12.0 $\pm$ 0.8 & 4.5 $\pm$ 0.5 \\
Real life cases & \textbf{84.1} $\pm$ 1.8 & 60.9 $\pm$ 2.2 & 63.9 $\pm$ 2.1 & 58.9 $\pm$ 2.3 & 76.6 $\pm$ 2.1 & 76.3 $\pm$ 1.9 \\
Hard bussiness cases & 69.1 $\pm$ 12.2 & 65.8 $\pm$ 13.1 & 67.9 $\pm$ 13.5 & 34.2 $\pm$ 11.1 & \textbf{69.6} $\pm$ 13.3 & 58.8 $\pm$ 13.5 \\
Agreements & \textbf{57.4} $\pm$ 3.2 & 35.3 $\pm$ 3.0 & 45.9 $\pm$ 3.1 & 32.6 $\pm$ 3.1 & 51.7 $\pm$ 3.0 & 44.5 $\pm$ 3.1 \\
GHEGA-based & 44.3 $\pm$ 4.9 & 45.1 $\pm$ 4.7 & \textbf{54.2} $\pm$ 4.6 & 32.2 $\pm$ 4.4 & 45.2 $\pm$ 4.9 & 30.7 $\pm$ 4.2 \\
Infographics-based & 73.0 $\pm$ 1.4 & 56.7 $\pm$ 1.6 & \textbf{74.3} $\pm$ 1.5 & 29.5 $\pm$ 1.4 & 67.3 $\pm$ 1.6 & 54.7 $\pm$ 1.6 \\
Financial & 50.8 $\pm$ 4.6 & 47.2 $\pm$ 4.4 & \textbf{52.6} $\pm$ 4.5 & 38.2 $\pm$ 4.1 & 46.3 $\pm$ 4.6 & 40.9 $\pm$ 4.2 \\
SEC-based & \textbf{92.7} $\pm$ 2.9 & 0.0 $\pm$ nan & 0.0 $\pm$ nan & 37.7 $\pm$ 9.8 & 63.7 $\pm$ 5.1 & 69.7 $\pm$ 2.7 \\
Signatures & \textbf{52.6} $\pm$ 4.4 & 30.7 $\pm$ 4.1 & 30.8 $\pm$ 4.0 & 28.0 $\pm$ 3.6 & 30.7 $\pm$ 4.1 & 28.5 $\pm$ 4.1 \\
Checkboxes & \textbf{72.8} $\pm$ 7.4 & 51.2 $\pm$ 8.6 & 60.7 $\pm$ 7.6 & 34.0 $\pm$ 7.3 & 59.9 $\pm$ 7.6 & 59.2 $\pm$ 8.2 \\
\midrule
\textbf{Average} & \textbf{66.5} & 39.7 & 46.6 & 35.1 & 52.3 & 46.8 \\
\bottomrule
\end{tabular}
\end{adjustbox}
\end{table*}

The error analysis reveals significant processing failures among cloud-based models. As shown in Table \ref{tab:error_rates_vision}, GPT5 and Claude 4 Sonnet experience substantial error rates across multiple datasets, with particularly high failure rates on the SEC-based dataset (100\% for both models), CUAD dataset (25\% for Claude), and Real life cases dataset (26.69\% for both). These failures likely stem from platform limitations including input size restrictions, memory constraints, and API timeouts when processing large business documents through the Cortex platform, rather than inherent model limitations.

Claude 4 Sonnet shows an average error rate of 20.64\% across all vision datasets, while GPT5 exhibits a 17.29\% average error rate. 

\begin{table}[H]
\centering
\caption{Error rates of GPT and Claude 4 Sonnet on internal vision datasets.}
\label{tab:error_rates_vision}
\begin{adjustbox}{center}
\begin{tabular}{@{}lcc@{}}
\toprule
\textbf{Dataset} & \textbf{GPT} & \textbf{Claude} \\
\midrule
CUAD & 5.95\% & 25\% \\
Real life cases & 26.69\% & 26.69\% \\
Hard bussiness cases & 0\% & 0\% \\
Agreements & 20.19\% & 26.63\% \\
GHEGA-based & 0\% & 0\% \\
Infographics-based & 0\% & 0\% \\
Financial & 0\% & 0\% \\
SEC-based & 100\% & 100\% \\
Signatures & 8.23\% & 13.37\% \\
Checkboxes & 11.86\% & 14.69\% \\
\midrule
\textbf{Average} & \textbf{17.29\%} & \textbf{20.64\%} \\
\bottomrule
\end{tabular}
\end{adjustbox}
\end{table}

\begin{landscape}
\begin{table*}[htbp]
\centering
\small
\caption{Performance comparison on multilingual datasets.}
\label{tab:multilingual_evaluation}
\begin{adjustbox}{center}
\renewcommand{\arraystretch}{1.2}
\begin{tabular}{@{}llcccccccccc@{}}
\toprule
\textbf{Dataset} & \textbf{Language} & \textbf{Arctic-Extract} & \textbf{Qwen2.5-VL} & \textbf{Claude 4 Sonnet} & \textbf{Pixtral} & \textbf{LLama 4 Scout} & \textbf{Llama 3.1 405B} & \textbf{Mistral 7B} & \textbf{GPT5} & \textbf{LLama 3.1 8B} & \textbf{TILT} \\ 
\midrule
JaQuAD & Japanese & 81.0 & 80.1 & 80.3 & 78.7 & 79.6 & 81.3 & 70.9 & \textbf{82.1} & 70.9 & 4.8 \\
KorQuAD & Korean & \textbf{79.8} & 79.1 & 77.2 & 77.9 & 78.9 & 78.9 & 73.0 & 70.6 & 73.0 & 5.7 \\
\midrule
\multirow{7}{*}{MLQA} & Arabic & 53.7 & 48.2 & 56.0 & 52.0 & 53.3 & \textbf{56.3} & 48.3 & 55.2 & 48.3 & 5.0 \\
 & German & 58.1 & 57.4 & 51.2 & \textbf{59.1} & 52.4 & 53.7 & 52.6 & 57.5 & 52.6 & 50.6 \\
 & English & \textbf{76.9} & 71.1 & 65.3 & 69.9 & 69.0 & 68.0 & 62.2 & 69.2 & 62.2 & 70.9 \\
 & Spanish & \textbf{65.6} & 56.5 & 57.9 & 59.6 & 58.7 & 62.0 & 59.1 & 62.6 & 59.1 & 56.7 \\
 & Hindi & 50.7 & 46.8 & 55.4 & 47.9 & 56.5 & 59.4 & 54.2 & \textbf{60.2} & 54.2 & 3.0 \\
 & Vietnamese & 53.4 & 53.8 & 53.8 & 53.6 & 54.6 & 57.5 & 52.5 & \textbf{58.5} & 52.5 & 14.8 \\
 & Chinese & \textbf{60.7} & 55.9 & 52.1 & 54.3 & 56.9 & 57.5 & 55.3 & 54.9 & 55.3 & 6.7 \\
\midrule
\multirow{11}{*}{xQuAD} & Arabic & \textbf{67.4} & 63.2 & 64.7 & 59.0 & 63.2 & 65.9 & 57.8 & 66.5 & 57.8 & 4.1 \\
 & German & \textbf{72.4} & 71.6 & 68.4 & 69.8 & 68.6 & 72.0 & 64.1 & 69.3 & 64.1 & 64.3 \\
 & Greek & \textbf{73.9} & 66.6 & 65.9 & 66.0 & 67.9 & 70.4 & 59.1 & 68.4 & 59.1 & 14.4 \\
 & English & \textbf{81.3} & 75.5 & 69.1 & 76.3 & 74.0 & 72.6 & 68.3 & 72.4 & 68.3 & 77.8 \\
 & Spanish & \textbf{74.8} & 67.8 & 65.7 & 68.7 & 66.7 & 70.6 & 61.9 & 68.2 & 61.9 & 68.3 \\
 & Hindi & 59.6 & 54.0 & 61.7 & 50.2 & 62.1 & \textbf{66.3} & 59.3 & 64.9 & 59.3 & 6.1 \\
 & Romanian & \textbf{74.8} & 69.5 & 67.3 & 67.8 & 67.2 & 70.3 & 64.4 & 69.1 & 64.4 & 52.6 \\
 & Russian & \textbf{69.8} & 65.5 & 62.7 & 62.3 & 65.8 & 65.0 & 58.4 & 67.0 & 58.4 & 22.2 \\
 & Thai & 72.3 & \textbf{72.7} & 69.3 & 53.7 & 68.9 & 72.0 & 62.1 & 68.0 & 62.1 & 4.5 \\
 & Turkish & 62.0 & 58.0 & 63.3 & 57.6 & 61.6 & 64.1 & 52.8 & \textbf{64.7} & 52.8 & 19.6 \\
 & Vietnamese & \textbf{66.6} & 63.6 & 63.2 & 59.3 & 61.8 & 66.4 & 58.3 & 64.6 & 58.3 & 17.5 \\
 & Chinese & \textbf{77.5} & 75.5 & 71.4 & 69.0 & 72.6 & 72.7 & 65.4 & 69.7 & 65.4 & 5.3 \\
\midrule
GermanQuAD-based & German & \textbf{84.1} & 64.4 & 76.2 & 53.6 & 70.8 & 77.1 & 8.7 & 79.5 & 58.9 & 71.7 \\
SQAC-based & Spanish & \textbf{79.7} & 60.9 & 64.6 & 49.5 & 63.8 & 66.6 & 5.0 & 68.0 & 53.6 & 66.5 \\
frenchQA-based & French & \textbf{90.5} & 83.5 & 86.3 & 35.5 & 82.3 & 84.1 & 10.1 & 85.6 & 45.4 & 89.4 \\
SQuAD-it-based & Italian & \textbf{72.1} & 50.3 & 64.7 & 45.6 & 54.5 & 62.9 & 1.1 & 63.5 & 52.5 & 63.5 \\
PoQuAD-based & Polish & \textbf{76.3} & 50.7 & 62.4 & 37.3 & 54.7 & 67.0 & 6.7 & 62.5 & 41.5 & 44.6 \\
BR-QuAD-2.0-based & Portuguese & \textbf{74.0} & 45.8 & 61.6 & 26.1 & 36.0 & 47.1 & 0.7 & 59.9 & 30.6 & 68.6 \\
\midrule
\multicolumn{2}{l}{\textbf{Average}} & \textbf{70.7} & 63.3 & 65.1 & 57.8 & 63.8 & 67.0 & 47.9 & 66.8 & 57.1 & 36.3 \\
\bottomrule
\end{tabular}
\end{adjustbox}
\end{table*}
\end{landscape}

\section{Challenges of Table Extraction}
\label{te_challenges}

\begin{figure}[H]
    \centering
    \includegraphics[width=0.7\textwidth]{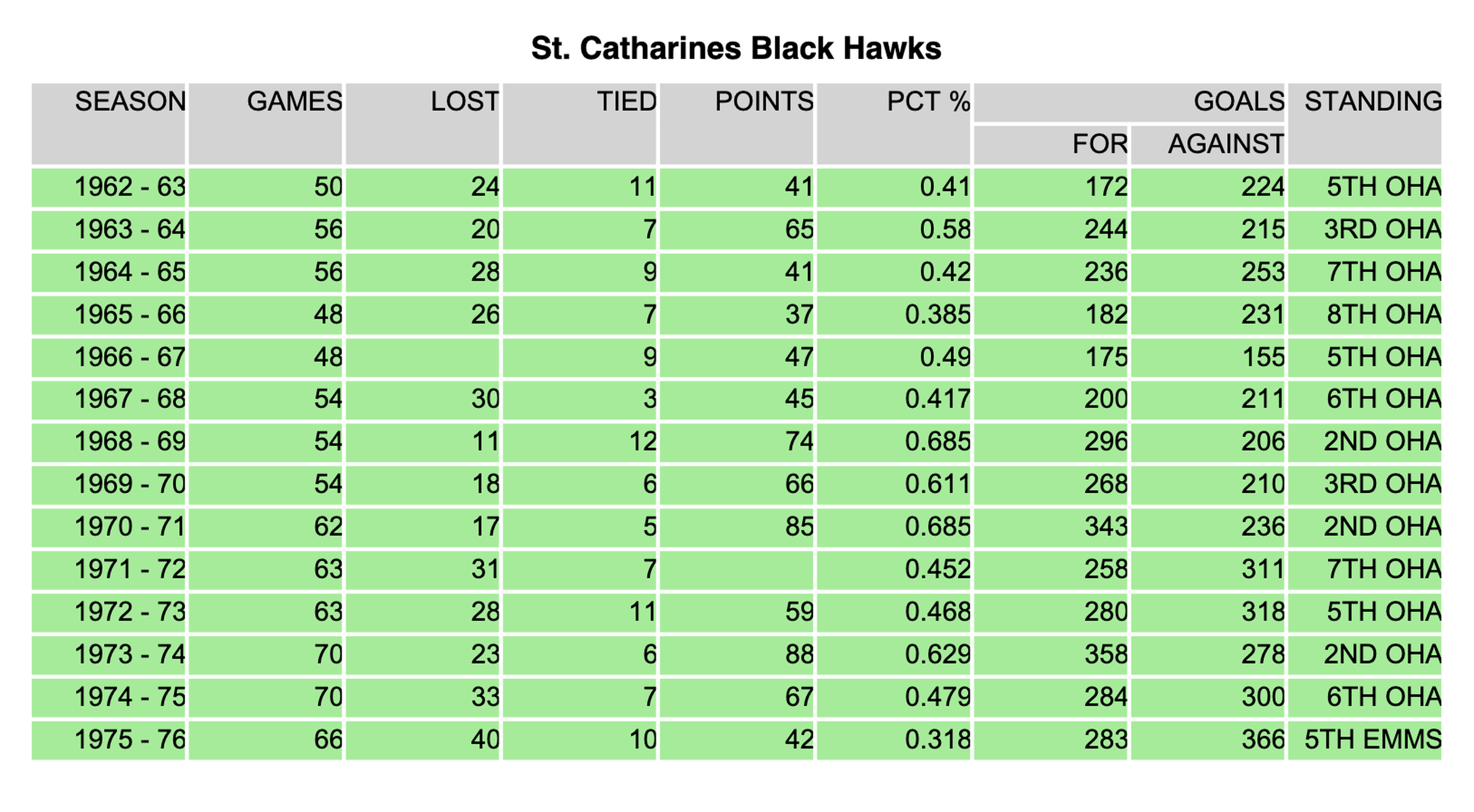} 
    \caption{Example of table with hierarchical header.}
    \label{fig:hierarchical_headers}
\end{figure}

\begin{figure}[H]
    \centering
    \includegraphics[width=0.7\textwidth]{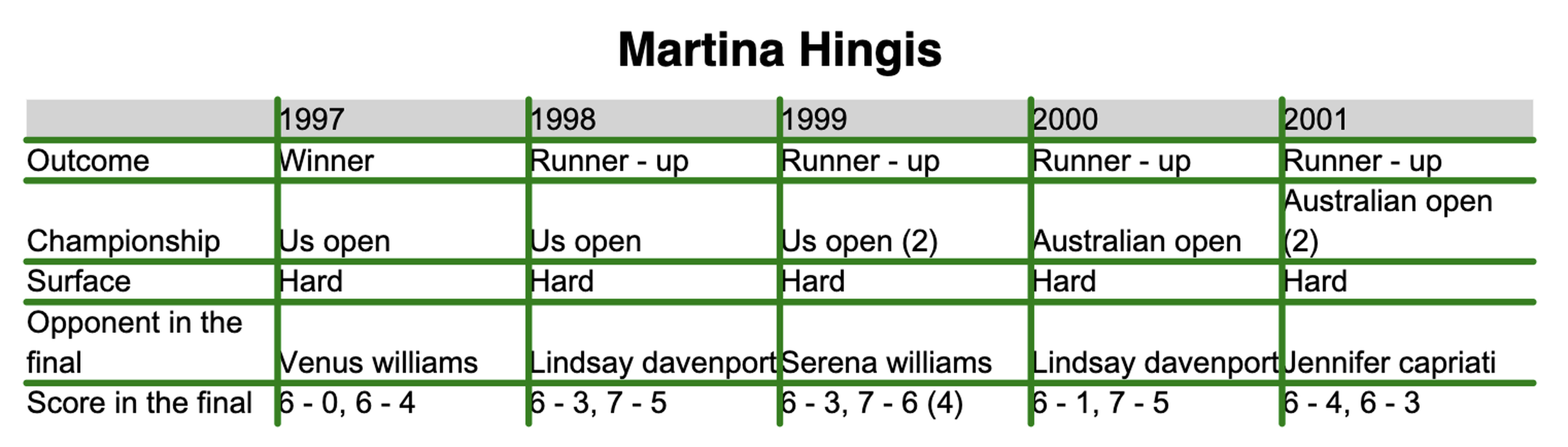} 
    \caption{Example of table transposition.}
    \label{fig:table_transposition}
\end{figure}

\begin{figure}[H]
    \centering
    \includegraphics[width=0.7\textwidth]{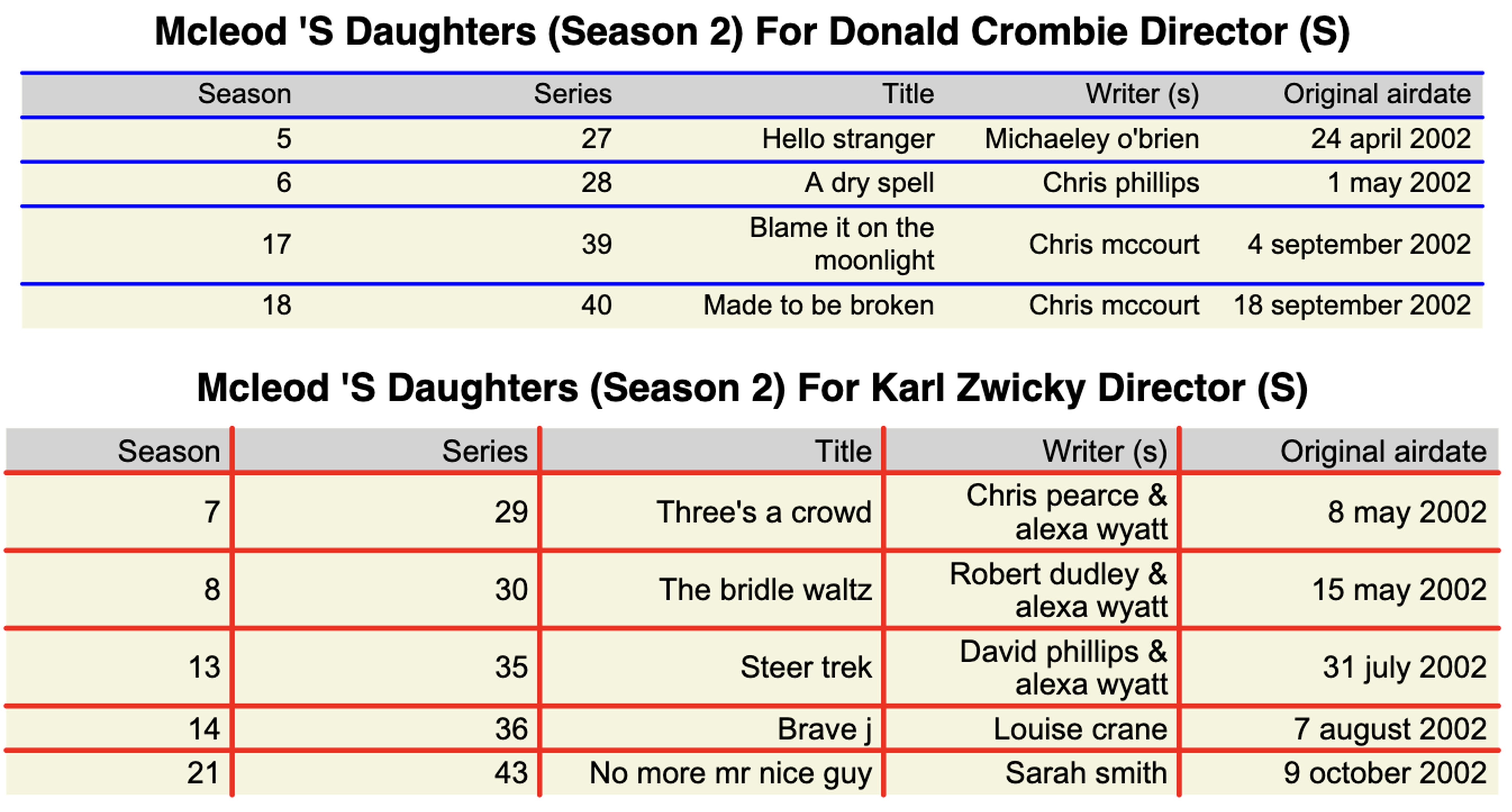} 
    \caption{Example of multiple table sources to join.}
    \label{fig:joining_multiple_sources}
\end{figure}

Table Extraction presents several unique challenges:
\begin{itemize}
  \itemsep0em 
  \item \textbf{Diverse Table Structures:} Tables can vary significantly in their layout, including presence of merged cells, repeated headers, table subsections (see example in Figure \ref{fig:table_subsections}) and hierarchical headers (see example in Figure \ref{fig:hierarchical_headers}).
  \item \textbf{Complex Formatting:} Tables often include complex formatting elements such as shading, varying font styles, and lack of borders. These elements can interfere with the accurate identification of table boundaries and cell contents.
  \item \textbf{Table Transposition:} In many document collections, it is more natural to represent data in a transposed format, where headers are presented in rows rather than columns. This requires the model to adapt to different orientations of data presentation (see example in Figure \ref{fig:table_transposition}).
  \item \textbf{Joining data from multiple sources:} In many cases, tables may span multiple pages or be split across different sections of a document. Extracting the complete table requires the ability to join data from these disparate sources accurately (see example in Figure \ref{fig:joining_multiple_sources}).
  \item \textbf{Repeating rows and empty cells:} Tables may contain repeating rows or empty cells, which can complicate the extraction process. Identifying and handling these cases is crucial for accurate data representation.
  \item \textbf{Contextual Understanding:} Extracting meaningful information from tables often requires understanding the context in which the data is presented. This includes recognizing headers, footnotes, and other contextual cues that provide additional meaning to the table contents. It also includes understanding the relationship between different tables within the same document.
\end{itemize}

All examples in this section are extracted from one of the training datasets based on TABFACT \cite{2019TabFactA}.

\subsection{Table Subsections}

\begin{figure}[H]
    \centering
    \includegraphics[width=0.8\textwidth]{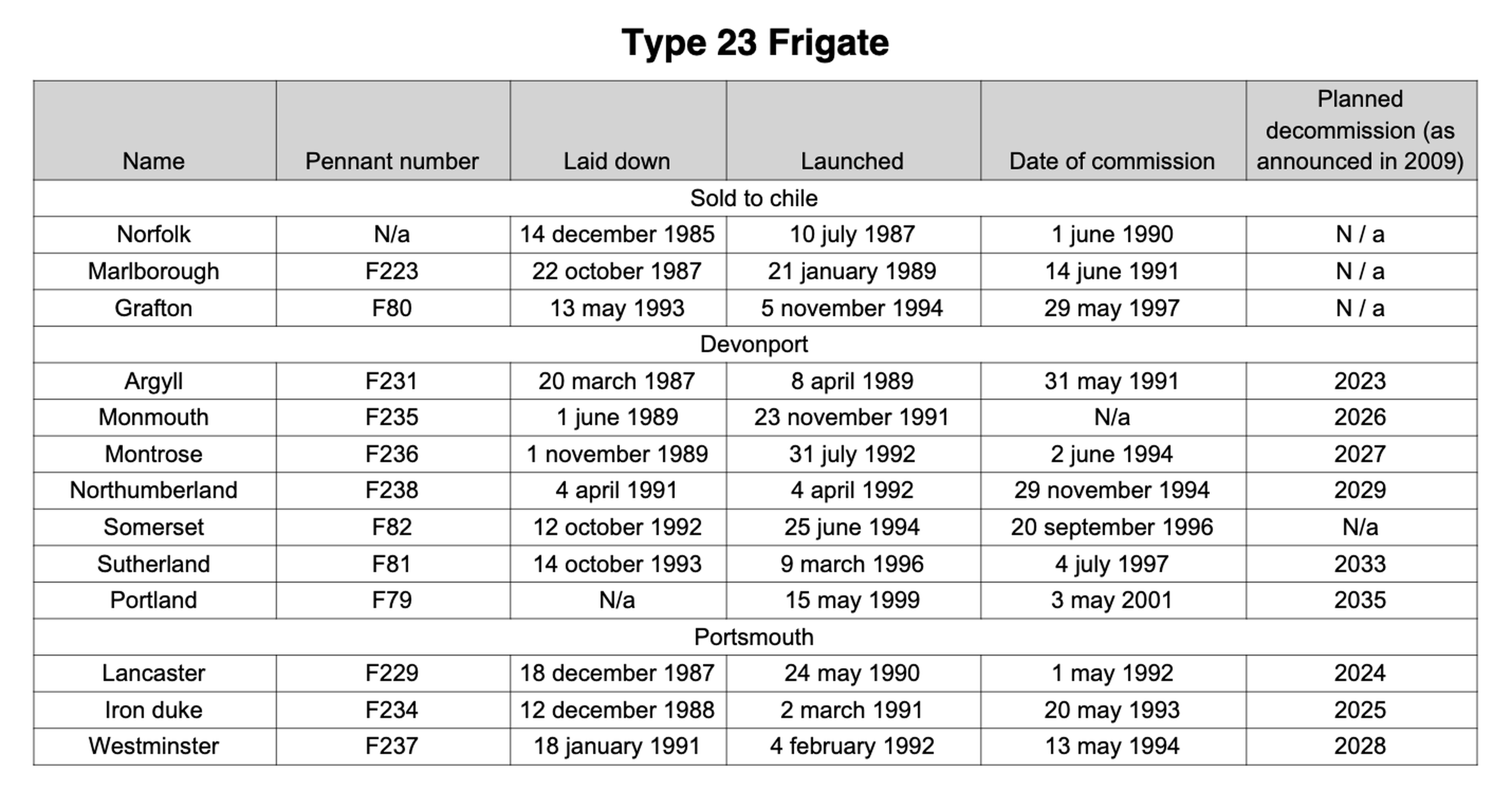} 
    \caption{Example of table with subsections.}
    \label{fig:table_subsections}
\end{figure}

When processing a table that contains subsections, each subsection can be represented as a distinct column in which the corresponding value repeats across all rows that belong to that subsection. The column name should be semantically meaningful and accurately describe the content of the subsection. In many cases, this name must be inferred from the surrounding context or domain knowledge.

For example, based on Figure \ref{fig:table_subsections}, a possible header for the extracted table could be:
\begin{displayquote}
\footnotesize
\texttt{Homeport | Name | Pennant number | Laid down}
\end{displayquote}

The expected model output (in rendered format) would be:

\begin{table}[H]
\centering
\caption{Extracted table with subsections.}
\begin{adjustbox}{center}
\small % Set smaller font size
\begin{tabular}{|l|l|l|l|l|l|}
\hline
\textbf{Homeport} & \textbf{Name} & \textbf{Pennant number} & \textbf{Laid down} \\ \hline
Sold to chile & Norfolk & N/a & 14 december 1985 \\ \hline
Sold to chile & Marlborough & F223 & 22 october 1987 \\ \hline
Sold to chile & Grafton & F80 & 13 may 1993\\ \hline
Devonport & Argyll & F231 & 20 march 1987  \\ \hline
Devonport & Monmouth & F235 & 1 june 1989  \\ \hline
Devonport & Montrose & F236 & 1 november 1989  \\ \hline
Devonport & Northumberland & F238 & 4 april 1991  \\ \hline
Devonport & Somerset & F82 & 12 october 1992  \\ \hline
Devonport & Sutherland & F81 & 14 october 1993  \\ \hline
Devonport & Portland & F79 & N/a  \\ \hline
Portsmouth &Lancaster & F229 & 18 december 1987  \\ \hline
Portsmouth &Iron duke & F234 & 12 december 1988  \\ \hline
Portsmouth &Westminster & F237 & 18 january 1991  \\ \hline
\end{tabular}
\end{adjustbox}
\end{table}

\subsection{Hierarchical Headers}

There are many cases where the source table header is hierarchical, meaning that some headers are grouped under broader categories. In such cases, the model should flatten the hierarchy by combining the broader category with the specific headers. In the example there is only one level of hierarchy. However, in more complex tables, there may be multiple levels of hierarchy that need to be flattened.

\begin{table}[H]
\centering
\caption{Extracted table with hierarchical header.}
\begin{adjustbox}{center}
\small % Set smaller font size
\begin{tabular}{|l|l|l|l|l|l|l|l|l|}
\hline
\textbf{SEASON} & \textbf{GAMES} & \textbf{LOST} & \textbf{TIED} & \textbf{POINTS} & \textbf{PCT \%} & \textbf{GOALS FOR} & \textbf{GOALS AGAINST} & \textbf{STANDING} \\ \hline
1962 - 63 & 50 & 24 & 11 & 41 & 0.41 & 172 & 224 & 5TH OHA \\ \hline
1963 - 64 & 56 & 20 & 7 & 65 & 0.58 & 244 & 215 & 3RD OHA \\ \hline
1964 - 65 & 56 & 28 & 9 & 41 & 0.42 & 236 & 253 & 7TH OHA \\ \hline
1965 - 66 & 48 & 26 & 7 & 37 & 0.385 & 182 & 231 & 8TH OHA \\ \hline
1966 - 67 & 48 & None & 9 & 47 & 0.49 & 175 & 155 & 5TH OHA \\ \hline
1967 - 68 & 54 & 30 & 3 & 45 & 0.417 & 200 & 211 & 6TH OHA \\ \hline
1968 - 69 & 54 & 11 & 12 & 74 & 0.685 & 296 & 206 & 2ND OHA \\ \hline
1969 - 70 & 54 & 18 & 6 & 66 & 0.611 & 268 & 210 & 3RD OHA \\ \hline
1970 - 71 & 62 & 17 & 5 & 85 & 0.685 & 343 & 236 & 2ND OHA \\ \hline
1971 - 72 & 63 & 31 & 7 & None & 0.452 & 258 & 311 & 7TH OHA \\ \hline
1972 - 73 & 63 & 28 & 11 & 59 & 0.468 & 280 & 318 & 5TH OHA \\ \hline
1973 - 74 & 70 & 23 & 6 & 88 & 0.629 & 358 & 278 & 2ND OHA \\ \hline
1974 - 75 & 70 & 33 & 7 & 67 & 0.479 & 284 & 300 & 6TH OHA \\ \hline
1975 - 76 & 66 & 40 & 10 & 42 & 0.318 & 283 & 366 & 5TH EMMS \\ \hline
\end{tabular}
\end{adjustbox}
\end{table}

\subsection{Table Transposition}

In this example, the source table presents data in a transposed format, where the headers are listed in the first column and the corresponding values are presented in subsequent columns. Column name "Year" is not explicitely mentioned as a header but can be derived from the values in first row. The model is expected to transpose the table during extraction, including "Year" column, resulting in a more conventional tabular format.

\begin{table}[H]
\centering
\caption{Extracted transposed table.}
\begin{adjustbox}{center}
\small % Set smaller font size
\begin{tabular}{|l|l|l|l|l|l|l|l|l|}
\hline
\textbf{Year} & \textbf{Outcome} & \textbf{Championship} & \textbf{Surface} & \textbf{Opponent in the final} & \textbf{Score in the final} \\ \hline
1997 & Winner & Us open & Hard & Venus williams & 6 - 0, 6 - 4 \\ \hline
1998 & Runner - up & Us open & Hard & Lindsay davenport & 6 - 3, 7 - 5 \\ \hline
1999 & Runner - up & Us open (2) & Hard & Serena williams & 6 - 3, 7 - 6 (4) \\ \hline
2000 & Runner - up & Australian open & Hard & Lindsay davenport & 6 - 1, 7 - 5 \\ \hline
2001 & Runner - up & Australian open (2) & Hard & Jennifer capriati & 6 - 4, 6 - 3 \\ \hline
\end{tabular}
\end{adjustbox}
\end{table}

\subsection{Joining Multiple Sources}

The model is expected to join data from different tables within the same document. Both tables in the example provide information about the series directed by different individuals. The model should combine these two sources to create a comprehensive view of the data, including directors.

\begin{table}[H]
\centering
\caption{Extracted table joined from two sources with additional "Director (S)" column.}
\begin{adjustbox}{center}
\small % Set smaller font size
\begin{tabular}{|l|l|l|l|l|l|l|l|l|}
\hline
\textbf{Season} & \textbf{Series} & \textbf{Title} & \textbf{Director (S)} & \textbf{Writer (s)} & \textbf{Original airdate} \\ \hline
5 & 27 & Hello stranger & Donald crombie & Michaeley o'brien & 24 april 2002 \\ \hline
6 & 28 & A dry spell & Donald crombie & Chris phillips & 1 may 2002 \\ \hline
17 & 39 & Blame it on the moonlight & Donald crombie & Chris mccourt & 4 september 2002 \\ \hline
18 & 40 & Made to be broken & Donald crombie & Chris mccourt & 18 september 2002 \\ \hline
7 & 29 & Three's a crowd & Karl zwicky & Chris pearce \& alexa wyatt & 8 may 2002 \\ \hline
8 & 30 & The bridle waltz & Karl zwicky & Robert dudley \& alexa wyatt & 15 may 2002 \\ \hline
13 & 35 & Steer trek & Karl zwicky & David phillips \& alexa wyatt & 31 july 2002 \\ \hline
14 & 36 & Brave j & Karl zwicky & Louise crane & 7 august 2002 \\ \hline
21 & 43 & No more mr nice guy & Karl zwicky & Sarah smith & 9 october 2002 \\ \hline
\end{tabular}
\end{adjustbox}
\end{table}

\section{Prompts}

We have used a prompt similar to the one below to evaluate all the models. The precise prompts depended on the specific dataset, and contained some internal optimisation.

\begin{displayquote}
\footnotesize
\texttt{Extract data from the context given the question. Do not output the question. Do not write a full sentence, just provide the resulting data. Extract data as is. Do not make any changes to the original data. If data is not present in the context output "None" as the answer to the question.
Question: \{question\}}
\end{displayquote}

\section{Contributions}
\textbf{MC} R\&D and experiments, model training, dataset selection, writing the manuscript, preparing figures and tables.\\

\textbf{JO} Table extraction problem modeling, preparation of the training datasets (analysis of data quality, filtering the data, fixing quality issues), describing Table Extraction problem with examples in manuscript.\\

\textbf{WJ} Project idea \& early-stage leadership, initial R\&D \& experiments, table extraction problem modelling.

\end{document}